\documentclass[conference]{IEEEtran}
\IEEEoverridecommandlockouts
\usepackage{amsmath,amssymb,amsfonts}
\usepackage{algorithmic}
\usepackage{graphicx}
\usepackage{textcomp}
\usepackage{xcolor}
\usepackage{booktabs}
\usepackage{makecell}
\usepackage{multirow}
\usepackage{subfigure}
\usepackage{subcaption}
\usepackage{hyperref}
\usepackage{amsthm} 
\usepackage{diagbox}
\newtheorem{theorem}{Theorem}

\usepackage{cite}

\begin{document}

\title{Uncertainty Quantification With Noise Injection in Neural Networks: A Bayesian Perspective
}

\author{\IEEEauthorblockN{Xueqiong Yuan}
\IEEEauthorblockA{\textit{Shenzhen International Graduate School} \\
\textit{Tsinghua University}\\
Shenzhen, China \\
xq-yuan22@mails.tsinghua.edu.cn}
\and
\IEEEauthorblockN{Jipeng Li}
\IEEEauthorblockA{\textit{Shenzhen International Graduate School} \\
\textit{Tsinghua University}\\
Shenzhen, China \\
lijipeng22@mails.tsinghua.edu.cn}
\and
\IEEEauthorblockN{Ercan Engin Kuruoglu}
\IEEEauthorblockA{\textit{Shenzhen International Graduate School} \\
\textit{Tsinghua University}\\
Shenzhen, China \\
kuruoglu@sz.tsinghua.edu.cn}
\thanks{This work is supported by Shenzhen Science and Technology Innovation Commission under Grant JCYJ20220530143002005, Tsinghua University SIGS Start-up fund under Grant QD2022024C, and Shenzhen Ubiquitous Data Enabling Key Lab under Grant ZDSYS20220527171406015.}
}

\maketitle
\begin{abstract}
    Model uncertainty quantification involves measuring and evaluating the uncertainty linked to a model's predictions, helping assess their reliability and confidence. Noise injection is a technique used to enhance the robustness of neural networks by introducing randomness. In this paper, we establish a connection between noise injection and uncertainty quantification from a Bayesian standpoint. We theoretically demonstrate that injecting noise into the weights of a neural network is equivalent to Bayesian inference on a deep Gaussian process. Consequently, we introduce a Monte Carlo Noise Injection (MCNI) method, which involves injecting noise into the parameters during training and performing multiple forward propagations during inference to estimate the uncertainty of the prediction. Through simulation and experiments on regression and classification tasks, our method demonstrates superior performance compared to the baseline model.
\end{abstract}


\section{Introduction}
Deep learning models have emerged as a powerful tool for addressing a wide array of tasks, encompassing pattern recognition, semantic segmentation, natural language processing, and more. In some high-risk applications, such as medicine and autonomous driving, handling model uncertainty is crucial. Model uncertainty refers to the uncertainty associated with the predictions made by a machine learning model. If the quantification of model uncertainty is achievable, the model can furnish not only its prediction but also an associated confidence level, thereby enhancing the reliability of its output. If a prediction by a model has high uncertainty, we can require the model not to make a decision but instead allow humans to make the decision manually.

Traditional neural networks (NNs) are deterministic and lack this functionality. By assigning distributions to weights rather than fixed values, Bayesian Neural Networks (BNNs) \cite{mackay1992practical} output predictive distributions instead of predictive values, making model uncertainty estimation possible. However, due to the large number of parameters, expensive computational resources, and slow convergence, applying BNNs to large-scale data poses significant challenges. In addition to improving optimization methods, efforts have been made to approximate BNNs using non-Bayesian models. For instance, theoretical findings have shown that dropout in neural networks serves as an approximation to Bayesian inference for deep Gaussian processes, leading to the proposal of Monte Carlo (MC) dropout methods for quantifying model uncertainty \cite{gal2016dropout}. Similarly, it has been shown that batch normalization and data augmentation have a comparable effect \cite{teye2018bayesian, jiang2022capturing}.

Noise injection, i.e., injecting additive or multiplicative random noise to certain components in the neural network, is an effective regularization technique to enhance generalization and robustness. The targets to which noise is injected could be input \cite{reed1992regularization,Rusak_2020,yuan2024robustness}, weights \cite{dropconnect2013, PNI2019, ho2009weight}, and activations \cite{dropout2014,ANP}. Noise injection fundamentally introduces randomness into neural networks, naturally leading us to model uncertainty. The noise injection into the weights is especially intuitive, which is also the primary focus of this paper, as it directly introduces randomness to the parameters of the neural network. Most of the existing research on noise injection focuses on the improvement of neural network prediction performance and less on its interpretation from a Bayesian perspective. While multiplicative noise injection into activations (dropout) \cite{gal2016dropout} has been studied concerning Bayesian models, noise injection into weights still lacks a Bayesian theoretical explanation.

In this paper, we establish a relationship between injecting noise into weights and quantifying uncertainty through a Bayesian perspective, and propose a method named Monte Carlo Noise Injection (MCNI) for uncertainty quantification. Our key contributions can be summarized as follows: 
\begin{itemize}
    \item We theoretically demonstrate that injecting noise into the weights of a neural network is equivalent to Bayesian inference on a deep Gaussian process, to which the BNN converges in the limit of infinite width.
    \item We propose a Monte Carlo Noise Injection (MCNI) technique, which includes injecting noise into the parameters during training and conducting multiple forward propagations during inference to assess the prediction's uncertainty.
    \item We evaluate the predictive distribution using the toy dataset and find that our method yields better results compared to the baseline.
    \item We evaluate predictive performance, calibration performance and selective performance on both regression and classification tasks, and our approach exceeds the baseline.
\end{itemize}

The paper is organized as follows. Section \ref{relatedwork} reviews current research on noise injection and uncertainty quantification. In Section \ref{proof}, we outline our proof, recovering the relationship between weight noise injection and Bayesian models, with details in  \hyperlink{https://drive.google.com/file/d/1LbvQtwV-pnGEKLGLc4DB5t7wbuV_2fRG/view?usp=sharing}{supplemental materials}\footnote{https://drive.google.com/file/d/1LbvQtwV-pnGEKLGLc4DB5t7wbuV\_2fRG/view?usp=sharing}. Sections \ref{exp1} and \ref{exp2} present experimental setup and results on the toy and real-world datasets. Finally, we discuss inference time, draw conclusions, and outline future work in Section \ref{conclusion}.

\section{Related work}\label{relatedwork}
\subsection{Weight Noise Injection}

Theoretical analysis conducted by \cite{DBLP:journals/neco/An96} demonstrates that weight noise can reduce the number of hidden units and encourage sigmoidal units in the output layer to exhibit saturation. \cite{ho2009weight} performs a heuristic analysis of the weight noise injection in MLPs and concludes that it can serve to regularize the magnitude of the hidden unit output, which is consistent with the results in \cite{DBLP:journals/neco/An96}. \cite{dropconnect2013} proposes dropconnect, which randomly sets the weights to zero and helps prevent overfitting. This is equivalent to injecting multiplicative Bernoulli noise into the weights. \cite{PNI2019} proposes Parametric Noise Injection (PNI), injecting layer-wise trainable Gaussian noise to various locations, including input, activation, and weights, to enhance the robustness of neural networks against adversarial attacks. Currently, research on weight noise injection primarily focuses on enhancing the generalization performance or robustness of predictions. It remains largely experimental, without connection to uncertainty quantification or Bayesian models.

\subsection{Model Uncertainty Quantification}

In the existing literature, Bayesian approximation and ensemble learning techniques are two of the most commonly used uncertainty quantification methods.

BNNs \cite{mackay1992practical} replace deterministic weights in traditional neural networks with distributions, allowing for the estimation of the model uncertainty. Due to the intractability of the posterior distribution, various MCMC-based and variational inference-based methods have been proposed to train BNNs, including Bayes by Backprop \cite{BBB2015} (BBB), Stochastic Gradient Langevin Dynamics \cite{welling2011bayesian} (SGLD), Stochastic Gradient MCMC \cite{SGMCMC2019} (SG-MCMC), etc. 

Because of the significant increase in model complexity, the computational cost of training BNNs is huge, making it difficult to scale to large dataset tasks. Uncertainty estimation with non-Bayesian methods, exemplified by ensemble learning, alleviates this problem. The idea is intuitive, averaging the outputs of multiple learners for prediction and computing variance for uncertainty estimation, e.g., Deep Ensembles \cite{DeepEnsemble2017}.

While the ensemble-based approach is simple and empirically proven to be effective, it has been criticized for not being linked to Bayesian methodology. Recently, the relationship and combination of the above two paradigms have attracted increasing interest. There is an emerging trend in research that involves approximating BNNs using non-Bayesian methods. It has been theoretically shown that dropout in neural networks serves as an approximation to Bayesian inference for deep Gaussian processes. Therefore, MC dropout, involving forward propagating several times during the testing stage and calculating the variance of output as uncertainty estimation is proposed \cite{gal2016dropout}. Similarly, it has been demonstrated that Batch Normalization \cite{teye2018bayesian} and data augmentation with Gaussian noise \cite{jiang2022capturing} also provide Bayesian approximation. \cite{AnchorEnsembling2020} proposes to modify the loss function of the maximum a posteriori (MAP) estimation by adding a random term to the loss function and training multiple networks, and it proves that this is equivalent to sampling from the Bayesian posterior distribution.

There are also studies that propose adding noise to weights or activations during the testing phase, followed by multiple forward propagations to obtain results with uncertainty estimates \cite{mehrtash2020pep,mi2022training}, which is similar to our idea. However, these studies are primarily empirical and lack theoretical proof from a Bayesian perspective. Unlike our approach, which adds noise during both training and testing for uncertainty estimation, they only introduce noise during testing. While this facilitates the use of pre-trained models, performance heavily relies on the chosen model. Our research has a distinct objective, which is training a robust model while simultaneously conducting uncertainty estimation. Additionally, their methods require empirical adjustment of noise levels, whereas our method is able to learn these levels during training.

\section{THEORETICAL PROOF}\label{proof}

\subsection{Background Material}
\subsubsection{Gaussian Processes}
Gaussian Processes (GPs) \cite{damianou2013deep} are probabilistic models that specify distributions across functions, enabling the flexible modeling of intricate data relationships. A GP is defined by its mean function and covariance function, denoted as $f(x)\sim \mathcal{GP}(m(x),K(x,x'))$. Here is a simple example of the application of GPs. Given a training dataset consisting of $N$ inputs $\{ \textbf x_1, \hdots, \textbf x_N \}$ and their corresponding outputs $\{\textbf y_1, \hdots, \textbf y_N\}$, we aim to estimate a function $\textbf y = \textbf f(\mathbf{x})$ that is likely to have generated our observations. Following the Bayesian approach we would put some prior distribution over the space of functions $p(\textbf f)$. This distribution reflects our initial belief regarding which functions are more probable and which are less probable. We then look for the posterior distribution over the space of functions given our dataset $(\textbf X, \textbf Y)$:
$$
p(\textbf f | \textbf X, \textbf Y) \propto p(\textbf Y | \textbf X, \textbf f) p(\textbf f).
$$
By modeling our distribution over the space of functions with a Gaussian process we can analytically evaluate its corresponding posterior in regression or approximate in the case of classification tasks. For regression, we place a joint Gaussian distribution over all function values,
\begin{align} \label{eq:generative_model_reg}
\textbf F | \textbf X &\sim \mathcal{N}(\textbf 0, \textbf K(\textbf X, \textbf X)), \notag \\
\textbf Y | \textbf F &\sim \mathcal{N}(\textbf F, \tau^{-1} \textbf I_N) ,
\end{align}
where $\textbf F_i = \textbf f (\textbf X_i)$ and $\textbf K(\textbf X, \textbf X)$ is the covariance function. For classification,
\begin{align*}
\textbf F | \textbf X &\sim \mathcal{N}(\textbf 0, \textbf K(\textbf X, \textbf X)),\notag \\
\textbf Y | \textbf F &\sim \mathcal{N}(\textbf F, 0 \cdot \textbf I_N) , \\
c_n | \textbf Y &\sim \text{Categorical}\left( \exp(y_{nd}) / \left(\sum_{d'} \exp(y_{nd'})\right) \right) \notag
\end{align*}
for $n = 1, ..., N$ with observed class label $c_n$. 
\cite{neal1996bayesian} reveals the equivalence between single-layer BNNs and GPs, while \cite{lee2018deep} extends it to multi-layer cases:
\begin{theorem}\label{theorem}
     A multi-layer fully-connected BNN with an i.i.d. prior over its parameters is equivalent to a Gaussian process (GP), in the limit of infinite network width. Specifically, let $x$ be the input of the neural network, $b$ be the bias term, $W_{ij}^l$ be the element at position $(i,j)$ of the weight matrix of layer $l$, and $\sigma$ be the non-linear activation function. If the $i$th component of the output of the $l$th-layer NN, $z_i^l$, is computed as, 
     $$z_i^l(x)=b_i^l+\sum_{j=1}^{N_l}W_{ij}^l y_j^l(x), y_j^l(x)=\sigma(z_j^{l-1}(x)),
     $$ 
     where $y_j^l(x)$ are i.i.d. for every $j$. Then, $z_i^l \sim \mathcal{GP}(0, K^l)$, where $K^l$ is the covariance function determined by $\sigma$. 
\end{theorem}
In other words, Gaussian processes are equivalent to infinitely wide BNNs. Therefore, GP is a powerful tool for Bayesian inference of neural networks. 
\subsubsection{Variational Inference}
For a general problem, where $\boldsymbol{\omega}$ represents the model parameters. Given data $\textbf X$ and $\textbf Y$, we need to find the posterior $p(\boldsymbol{\omega}|\textbf X, \textbf Y)$. When the posterior distribution is intractable to solve, some approximation techniques need to be used, such as variational inference. Using variational inference, we introduce a variational distribution $q(\boldsymbol{\omega})$ with a simple property to approximate $p(\boldsymbol{\omega}|\textbf X, \textbf Y)$. The goal is to minimize the KL divergence between them: $KL(q(\boldsymbol{\omega})|p(\boldsymbol{\omega}|\textbf X, \textbf Y))$, which is equivalent to minimizing the negative log Evidence Lower Bound (ELBO) \cite{Bishop_2006_paper}:
\begin{equation*}
    \mathcal{L}_{\text{VI}} := -\int q(\boldsymbol{\omega}) \log p(\textbf Y | \textbf X, \boldsymbol{\omega}) \text d \boldsymbol{\omega} + \text{KL}(q(\boldsymbol{\omega}) || p(\boldsymbol{\omega})). \label{eq:L:VI}
\end{equation*}
\subsection{Weight Noise Injection As a Bayesian Approximation}
We theoretically prove that weight noise injection serves as a Bayesian approximation to the deep Gaussian process model. Let $\hat{\textbf{y}}$ be the output and $\textbf{x}$ be the input of a neural network respectively, with dimensions $D$ and $Q$. We start the discussion from a single-layer NN. Assuming there are $K$ units in the hidden layer. Let $\textbf{W}_1$ be the weight matrix of dimensions $Q \times K$, connecting the input layer and the hidden layer, and $\textbf{W}_2$ be the weight matrix of dimensions $K \times D$, connecting the hidden layer and the output layer. Let $\textbf{b}$ be a $K$ dimensional vector, and $\sigma(\cdot)$ be the element-wise non-linearity such as ReLU or tanh. Thus, a standard neural network can be represented as:
\begin{equation*}
    \hat{\textbf{y}}=\sigma(\textbf{x}\textbf{W}_1+\textbf{b})\textbf{W}_2.
\end{equation*}
Now consider a neural network injected with Gaussian noise for the weights. Denote $Z_1$ and $Z_2$ as matrices of $Q\times K$ and $K\times D$ respectively, of which the element $z_{1,i,j}\sim \mathcal{N}(0, \sigma_1^2)$ and $z_{2,i,j}\sim \mathcal{N}(0, \sigma_2^2)$, then this neural network can be represented as:
\begin{equation}\label{output_ninn}
    \hat{\textbf{y}}=\sigma(\textbf{x}(\textbf{W}_1+\textbf{Z}_1)+\textbf{b})(\textbf{W}_2+\textbf{Z}_2).
\end{equation}
Introducing weight decay, the loss function of this neural network is:
\begin{equation} \label{L_NN_NI}
    \mathcal{L}_{\text{NN-NI}}:=l(\hat{\textbf{y}}, \textbf{y})+\lambda_1 ||\textbf{W}_1||_2^2 +\lambda_2 ||\textbf{W}_2||_2^2 +\lambda_3 ||\textbf{b}||_2^2,
\end{equation}
where $l(\hat{\textbf{y}}, \textbf{y})$ is the mean squared error for regression tasks and cross entropy for classification tasks. Herein, we only discuss regression tasks, however, the extension to classification tasks can be easily achieved.

We are able to show that minimizing $\mathcal{L}_{\text{NN-NI}}$ in (\ref{L_NN_NI}) is equivalent to maximizing the ELBO of a particular Gaussian process. For the Gaussian process in (\ref{eq:generative_model_reg}), define the covariance function $\textbf{K}(\textbf{x}, \textbf{y}) = \int p(\textbf{w}) p(b) \sigma(\textbf{w}^T \textbf{x} + b) \sigma(\textbf{w}^T \textbf{y} + b) \text{d} \textbf{w} \text{d} \textbf{b}$ with $p(\textbf{w})$ as a standard multivariate normal distribution of dimensionality $Q$ and some distribution $p(b)$. Using Monte Carlo integration with $K$ terms to approximate the integral above: $\hat{\textbf{K}}(\textbf{x}, \textbf{y}) = 
\frac{1}{K} \sum_{k=1}^K 
 \sigma(\textbf{w}_k^T \textbf{x} + b_k) \sigma(\textbf{w}_k^T \textbf{y} + b_k)$ with $\textbf{w}_k \sim p(\textbf{w})$ and $b_k \sim p(b)$, where $K$ is the number of hidden units in our single hidden layer NN approximation, and replacing the variational distribution $q(\textbf{W}_1, \textbf{W}_2, \textbf{b}):=q(\textbf{W}_1)q(\textbf{W}_2)q(\textbf{b})$:
\begin{align*}
    q(\textbf{W}_1) &= \prod_{q=1}^Q q(\textbf{w}_q),\\
    q(\textbf{w}_q) &= \mathcal{N}(\textbf{m}_q, \boldsymbol{\sigma}^2 \textbf{I}_K) + \mathcal{N}(0, \boldsymbol{\sigma}_1^2 \textbf{I}_K), \\
    q(\textbf{W}_2) &= \prod_{k=1}^K q(\textbf{w}_k), \\
    q(\textbf{w}_k) &= \mathcal{N}(\textbf{m}_k, \boldsymbol{\sigma}^2 \textbf{I}_D) + \mathcal{N}(0, \boldsymbol{\sigma}_2^2 \textbf{I}_D), \\
    q(\boldsymbol{b}) &= \mathcal{N}(\textbf{m}, \boldsymbol{\sigma}^2 \textbf{I}_K),
\end{align*}
we can prove that the negative ELBO of variational inference is:
\begin{align*}
    \mathcal{L}_{\text{GP-MC}} &\propto  \frac{1}{2N} \sum_{n=1}^N || \textbf{y}_n - \hat{\textbf{y}}_n ||^2_2 + \frac{1}{2\tau N} || \textbf{M}_1||_2^2 \\ &+ \frac{1}{2\tau N} ||\textbf{M}_2||_2^2 + \frac{1}{2\tau N} ||\textbf{m}||_2^2,
\end{align*}
where $\textbf{M}_1 = [\textbf{m}_q]_{q=1}^{Q}$, $\textbf{M}_2 = [\textbf{m}_k]_{k=1}^K$, $N$ is the number of inputs, $\tau$ is the model precision and $\hat{\textbf{y}}_n$ is the output of a NN with weight noise injection, which is defined in (\ref{output_ninn}). Details of the proof are provided in supplemental materials. Therefore, minimizing $\mathcal{L}_{\text{GP-MC}}$ is equivalent to minimizing $\mathcal{L}_{\text{NN-NI}}$ in (\ref{L_NN_NI}). The above proof can be generalized to multi-layer neural networks, which is shown in supplemental materials. Thus, a neural network with weight noise injection could be approximated by a GP model, and according to Theorem \ref{theorem}, GP models have equivalence to BNNs. Accordingly, the predictive posterior distribution is:
\begin{equation*}
    q(\textbf{y}^*|\textbf{x}^*)=\int p(\textbf{y}^*|\textbf{x}^*,\boldsymbol{\omega})q(\boldsymbol{\omega})\text{d}\boldsymbol{\omega}
\end{equation*}
where $\boldsymbol{\omega}=\{\Tilde{\textbf{W}}_l\}_{l=1}^L$ is the noisy weights of neural network, and
\begin{align}\label{noisy_eq}
    &\Tilde{\textbf{W}}_l=\{\tilde{\textbf{w}}_{l,i,j}\}, \notag \\
    &\tilde{\textbf{w}}_{l,i,j}=\textbf{w}_{l,i,j}+\alpha_{l,i,j}\cdot \epsilon_{l,i,j}, \epsilon_{l,i,j} \sim \mathcal{N}(0, \sigma_l^2).
\end{align}
$\textbf{w}_{l,i,j}$ is element of noise-free weight matrix in the $l$-th layer of NN, $\sigma_l^2$ is the variance of weights in the $l$-th layer, and $\alpha_{l,i,j}$ is the coefficient of noise level. For the value of $\alpha_{l,i,j}$, we propose two versions: setting a fixed value and setting it as a learnable parameter. When it is a learnable parameter, it is optimized by the optimizer in backpropagation, like $w_{l,i,j}$. To prevent the learned $\alpha_{l,i,j}$ from becoming too small and approaching a deterministic NN, a penalty term $-\lambda||\alpha||_2^2$, where $\lambda>0$ can be introduced in the function as needed.

We simply use the Monte Carlo method to estimate the mean and variance of the predictive distribution:
\begin{align*}
    \mathbb{E}_{q(\textbf{y}^*|\textbf{x}^*)}(\textbf{y}^*) &\approx \frac{1}{T} \sum_{t=1}^T \hat{\textbf{y}}^* (\textbf{x}^*, \boldsymbol{\omega}),  \\
    \text{Var}_{q(\textbf{y}^*|\textbf{x}^*)}(\textbf{y}^*) &\approx \frac{1}{T-1} \sum_{t=1}^T (\hat{\textbf{y}}^* (\textbf{x}^*, \boldsymbol{\omega}) - \mathbb{E}_{q(\textbf{y}^*|\textbf{x}^*)}(\textbf{y}^*))^2.
\end{align*}
In practical terms, this entails conducting T forward passes during inference, computing the mean of the predicted values as the prediction, and calculating the variance of the predicted values as an estimate of uncertainty. We name it Monte Carlo Noise Injection (MCNI). In the following, we will refer to the methods where the noise level $\alpha$ is set to a fixed value or treated as a learnable parameter as MCNI (fixed) and MCNI (learned), respectively.

\section{Experiments on Toy Dataset}\label{exp1}

\subsection{Experiment Configuration}
To evaluate the predictive posterior distribution obtained by our proposed method, we generate the toy dataset through random simulation. We uniformly select 200 values as $x$ from $[-2, 2]$ and generate sample points using $y=0.3\sin \pi x + 0.2\epsilon$, where $\epsilon \sim \mathcal{N}(0, x^2)$. A single-layer neural network with 100 units is used to analyze the samples. Both MC Dropout and MCNI models are trained using the Adam optimizer with a learning rate of 0.005, 500 epochs, and a momentum of 0.9. For both methods, we perform 500 forward passes for estimation. 

\subsection{Metrics}
We use Prediction Interval Coverage Probability (PICP) and Mean Prediction Interval Width (MPIW) as metrics to evaluate the performance of the model and compare it with MC Dropout as the baseline. The formulas for PICP and MPIW are as follows:
\begin{align*}
    \text{PICP} &= \frac{1}{N} \sum_{i=1}^{N} \mathbb{I}\left(y_i \in \left[\hat{y}_i^\text{L}, \hat{y}_i^\text{U}\right]\right),\\
    \text{MPIW} &= \frac{1}{N} \sum_{i=1}^{N} \left(\hat{y}_i^\text{U} - \hat{y}_i^\text{L}\right),
\end{align*}
where $\hat{y}_i^\text{U}$ and $\hat{y}_i^\text{L}$ are the upper and lower bounds of the prediction interval for the $i$-th data point, and $N$ is the number of samples. Let $\hat{y}$ be the predicted mean and $\hat{\sigma}_y$ be the predicted standard deviation. We define the upper and lower bounds of the prediction interval as $\hat{y}^U = \hat{y} + 3\hat{\sigma}_y$ and $\hat{y}^L = \hat{y} - 3\hat{\sigma}_y$, respectively. 

\subsection{Result}
\begin{figure*}[h]
\centering
\subfigure[MC dropout, MPIW=0.4616, PICP=0.58]
{\begin{minipage}[b]{0.32\linewidth}
    \centering
    {\includegraphics[width=1\linewidth]{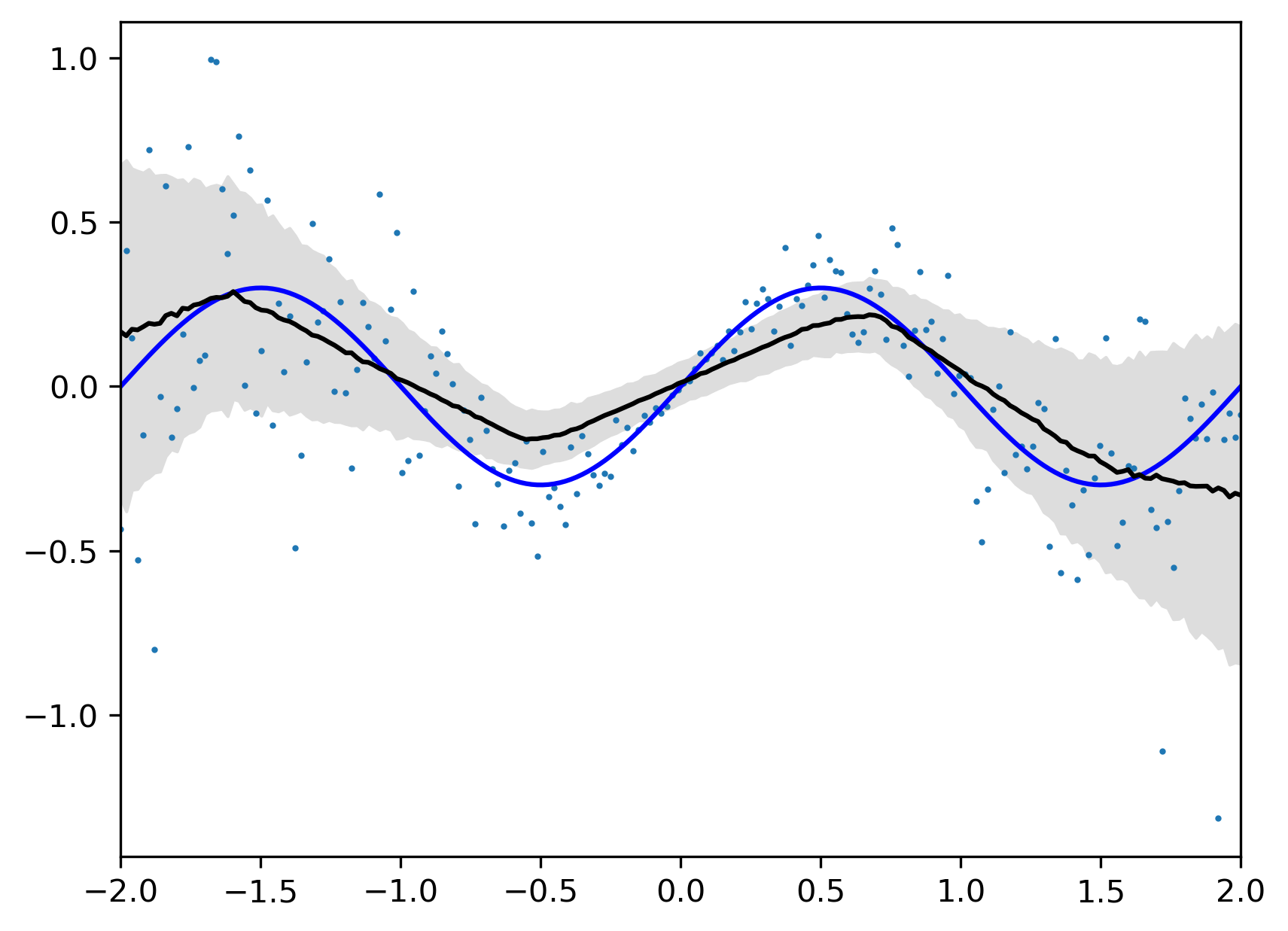}}
\end{minipage}
}
\subfigure[MCNI (fixed), MPIW=0.2915, PICP=0.625]
{\begin{minipage}[b]{0.32\linewidth}
    \centering
    {\includegraphics[width=1\linewidth]{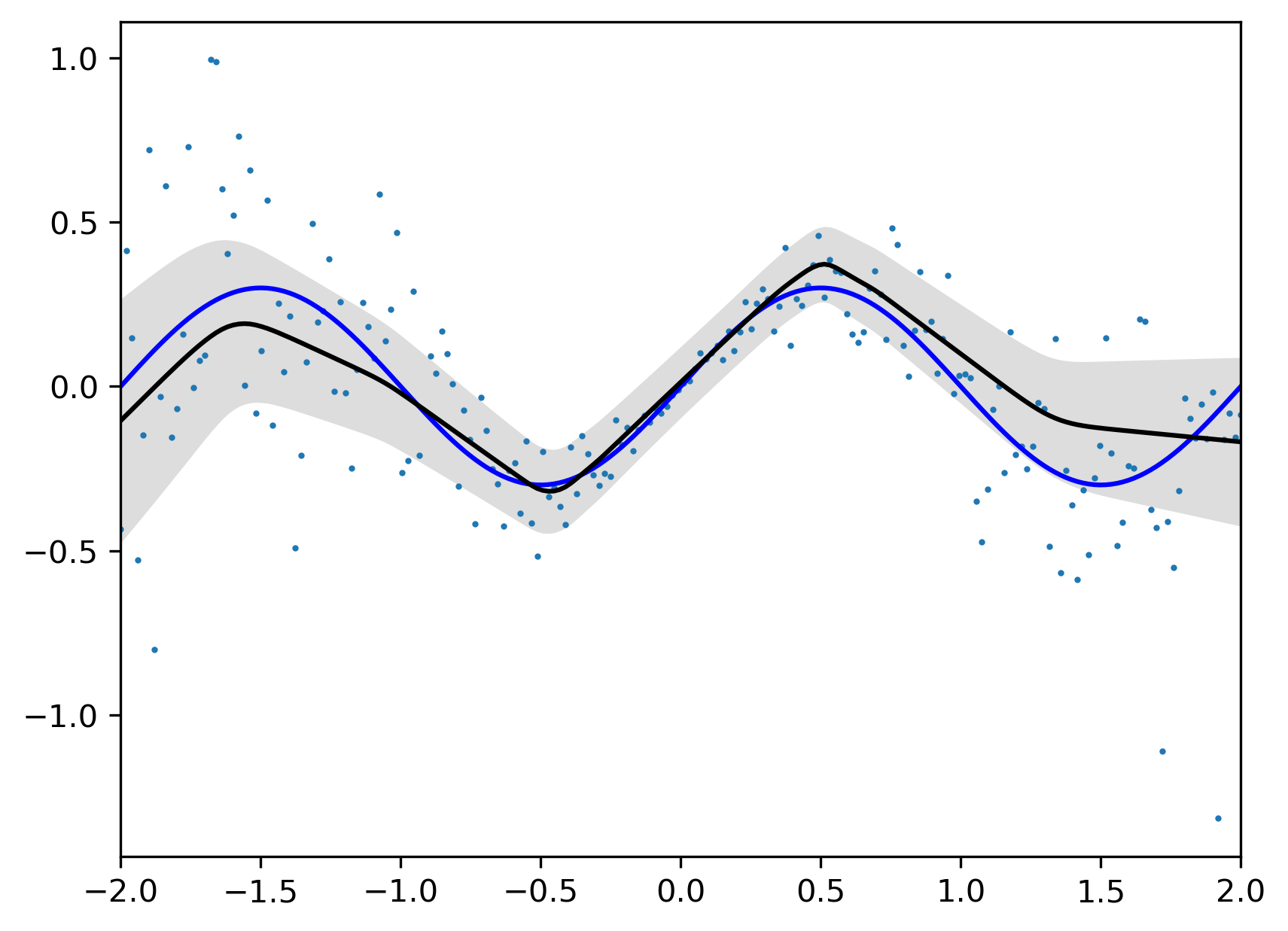}}
\end{minipage}
}
\subfigure[MCNI (learned), MPIW=0.3043, PICP=0.64]
{\begin{minipage}[b]{0.32\linewidth}
    \centering
    {\includegraphics[width=1\linewidth]{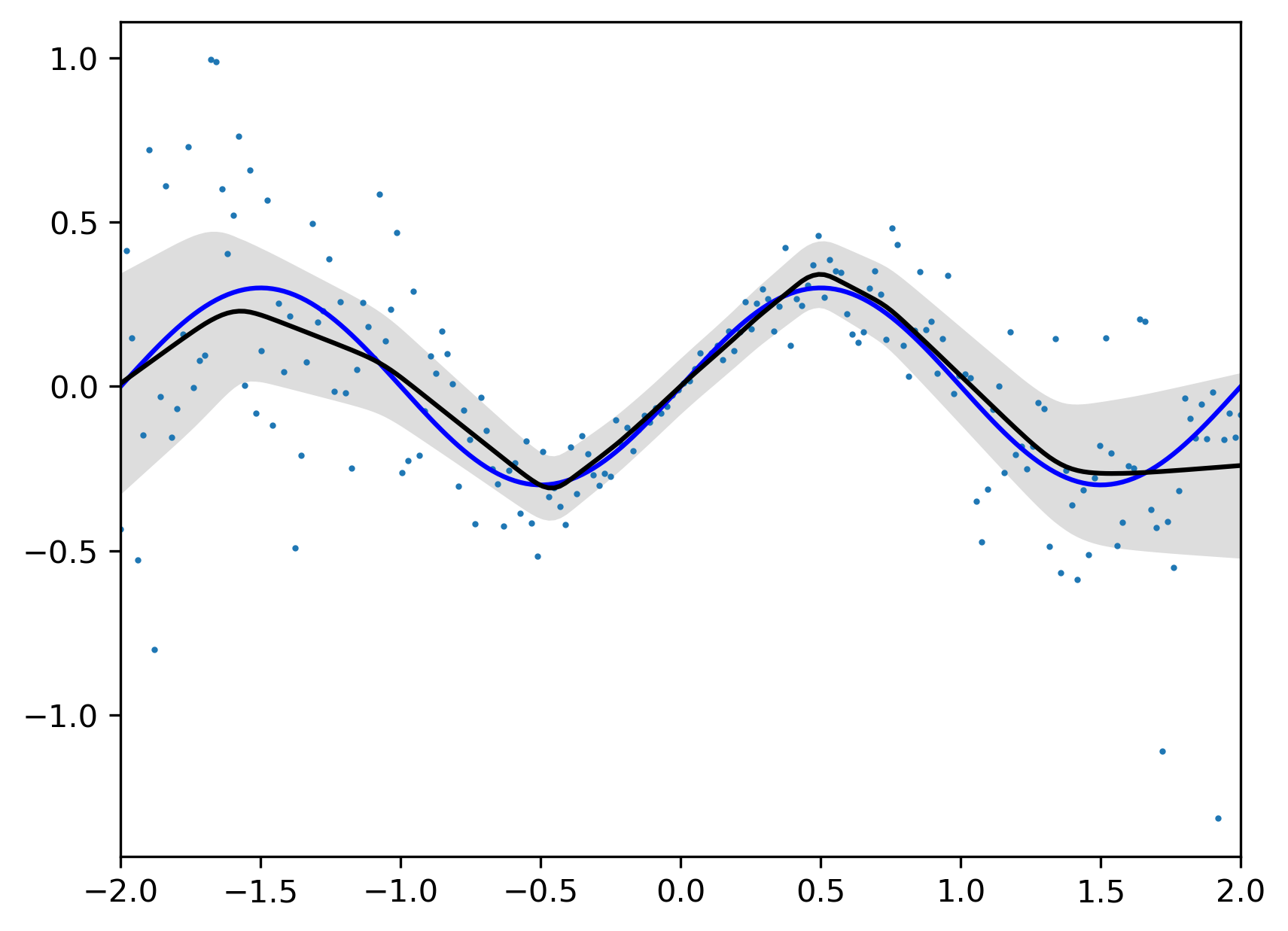}}
\end{minipage}
}
\caption{Predictions made by each method on the toy dataset. The data points are represented by light blue points, the true data generating function $y=0.3\sin \pi x$ is depicted as the dark blue line and the average predictions are shown as a black line. Shaded areas represent regions of predicted means $\pm 3$ standard deviations.}
\label{simulation}
\end{figure*}

By adjusting the dropout probability in MC Dropout or the noise level in MCNI, we can trade off between MPIW and PICP. However, we find that MCNI consistently provides narrower intervals with higher coverage, resulting in a smaller MPIW and a larger PICP compared to MC Dropout. Here, we present one set of results, where the dropout probability for MC Dropout is 0.2 and the $\alpha$ for MCNI (fixed) is 0.05. As shown in Fig. \ref{simulation}, the performance of MCNI (fixed) and MCNI (learned) is similar, both achieving smaller MPIW and larger PICP than MC dropout. Additionally, the predicted mean (black line) is closer to the ground truth (dark blue line), which indicates that MCNI provides more accurate interval estimates.

\section{Experiments on Real-World Datasets}\label{exp2}
Our methods are applied to the UCI regression dataset \cite{asuncion2007uci} and the CIFAR10 image dataset \cite{Krizhevsky09learningmultiple} to evaluate the predictive performance, calibration performance and selective performance. Furthermore, we visually depict the prediction distributions by showcasing examples from the CIFAR10 dataset.

\subsection{Datasets and Models}\label{dataset_model}
We use 8 classical regression datasets from the UCI Machine Learning Repository. For the choice of hyperparameters we refer to \cite{hernandez2015probabilistic}. Specifically, we utilize a two-layer neural network with 50 units, except for the protein structure dataset, where we employ a two-layer neural network with 100 units. With a batch size of 32, an Adam optimizer with a momentum of 0.9 is used. The maximum number of epochs is 2000 and an early stopping with patience = 3 is used. We conduct a grid search for the other hyperparameters, including learning rate, weight decay, dropout probability for MC dropout, noise level for MCNI (fixed), and initial value of learnable noise level for MCNI (learned) as listed in Table \ref{hyper}. We employ cross-validation to identify the set of parameters that yields the minimum validation loss for each model.

\begin{table*}[tb]
\caption{Hyperparameter grid search ranges for regression tasks.}
\label{hyper}
\begin{center}
\begin{tabular}{cc}
   \toprule
   Hyperparameter & Ranges \\
   \midrule
    Learning rate & [0.0001, 0.0005, 0.001, 0.002] \\
    Weight decay & [0.1, 0.01, 0.001, 1e-4, 1e-5, 1e-9] \\
    Dropout probability for MC dropout &  [0.001, 0.005, 0.01, 0.05, 0.1, 0.2] \\
    Noise level for MCNI (fixed) & [0.001, 0.005, 0.01, 0.05, 0.1]  \\
    Initial value of noise level for MCNI (learned) & [0.001, 0.005, 0.01, 0.05, 0.1]  \\
   \bottomrule
\end{tabular}
\end{center}
\end{table*}

CIFAR10 is a dataset of 60,000 $32\times32$ color images with 10 different categories, with 50,000 for training and 10,000 for testing. The model employs the architecture of ResNet8 \cite{he2016deep}, with a batch size of 128, utilizing stochastic gradient descent with a learning rate of 0.1, momentum of 0.9, and weight decay of 0.0003. The model is trained for 160 epochs. The noise level for MCNI (fixed) is 0.02. The dropout probability for MC dropout is set to 0.3 \cite{jiang2022capturing}. We also try dropout probabilities of 0.1, 0.2, and 0.5 and find the results are insensitive to it.

All the models are trained on a single NVIDIA RTX 6000 GPU.

\subsection{Metrics}
 For regression tasks, we use root mean squared error (RMSE) and mean standardized log loss (MSLL). MSLL is defined as the difference from the negative log-likelihood ($-\log {p(\hat{y}|X,y,\hat{x})} = \frac{1}{2} \log (2\pi \hat{\sigma}_y^2) + \frac{(y-\hat{y})^2}{2\hat{\sigma}_y^2}$) of the baseline model (MC dropout). Therefore, the MSLL of MC dropout is always 0. For classification tasks, we use the accuracy, the expected calibration error (ECE) \cite{guo2017calibration}, and the Brier score \cite{brier1950verification}. Among these, RMSE and accuracy measure predictive performance, while MSLL, ECE, and Brier score assess calibration performance, which are crucial indicators of uncertainty quality.
 
 In addition, we use risk-coverage curves to demonstrate the model's ability to select high-risk test samples. A risk-coverage curve is a graphical representation that illustrates the relationship between coverage (the proportion of samples for which the model's predicted confidence exceeds a given threshold) and risk (the misclassification error, or other risk metric, for predictions where the model's predicted confidence surpasses the threshold) in a predictive model. 
 
The implementation involves sorting all predictions for test samples in ascending order based on their uncertainty estimates. For different coverage rate  $x\%$ ranging from 0\% to 100\%, we select the top $x\%$ samples with the lowest predicted variance, i.e., the highest confidence level, and calculate their associated risk measures. This approach allows us to plot the risk-coverage curve. In a model capable of providing reliable uncertainty estimates, samples with lower uncertainty estimates should correspond to more accurate predictions. Consequently, as the coverage rate increases, risk is expected to also increase, reflecting the model's ability to effectively select high-risk samples. Specifically, risk is typically measured using a loss function or accuracy. In this study, we use RMSE for regression tasks and accuracy for classification tasks.

\begin{figure*}[htbp]
\centering
\subfigure[Vanilla]
{\begin{minipage}[b]{0.45\linewidth}
    \centering
    {\includegraphics[width=1\linewidth]{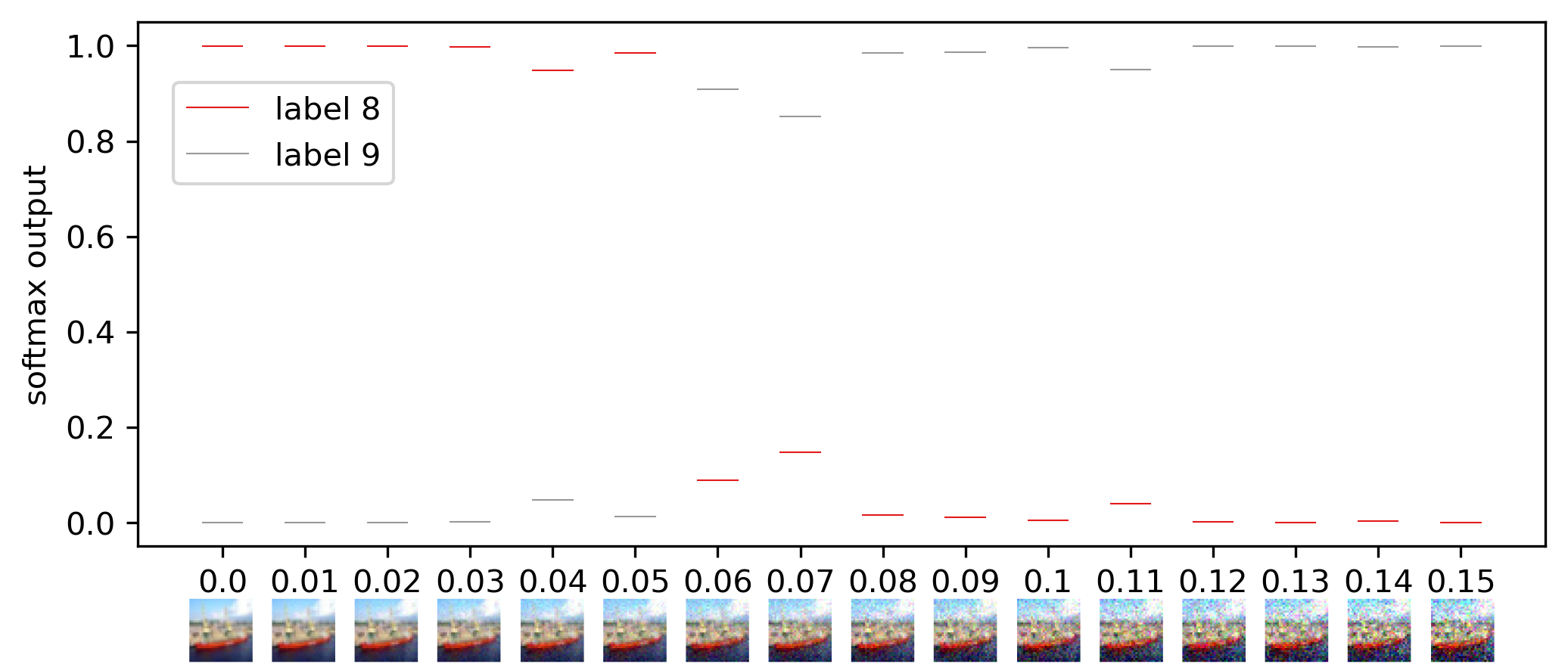}}
\end{minipage}
}
\subfigure[MC dropout]
{\begin{minipage}[b]{0.45\linewidth}
    \centering
    {\includegraphics[width=1\linewidth]{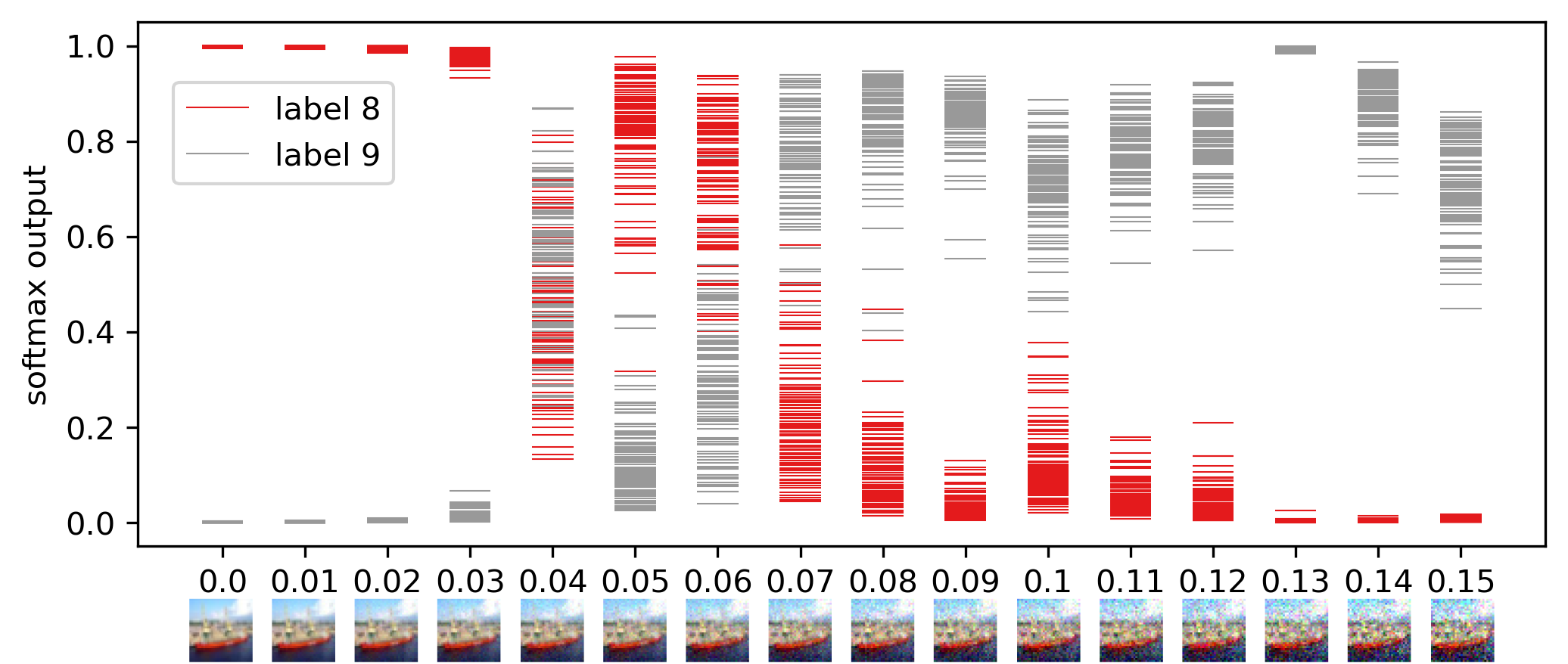}}
\end{minipage}
}
\subfigure[MCNI (fixed)]
{\begin{minipage}[b]{0.45\linewidth}
    \centering
    {\includegraphics[width=1\linewidth]{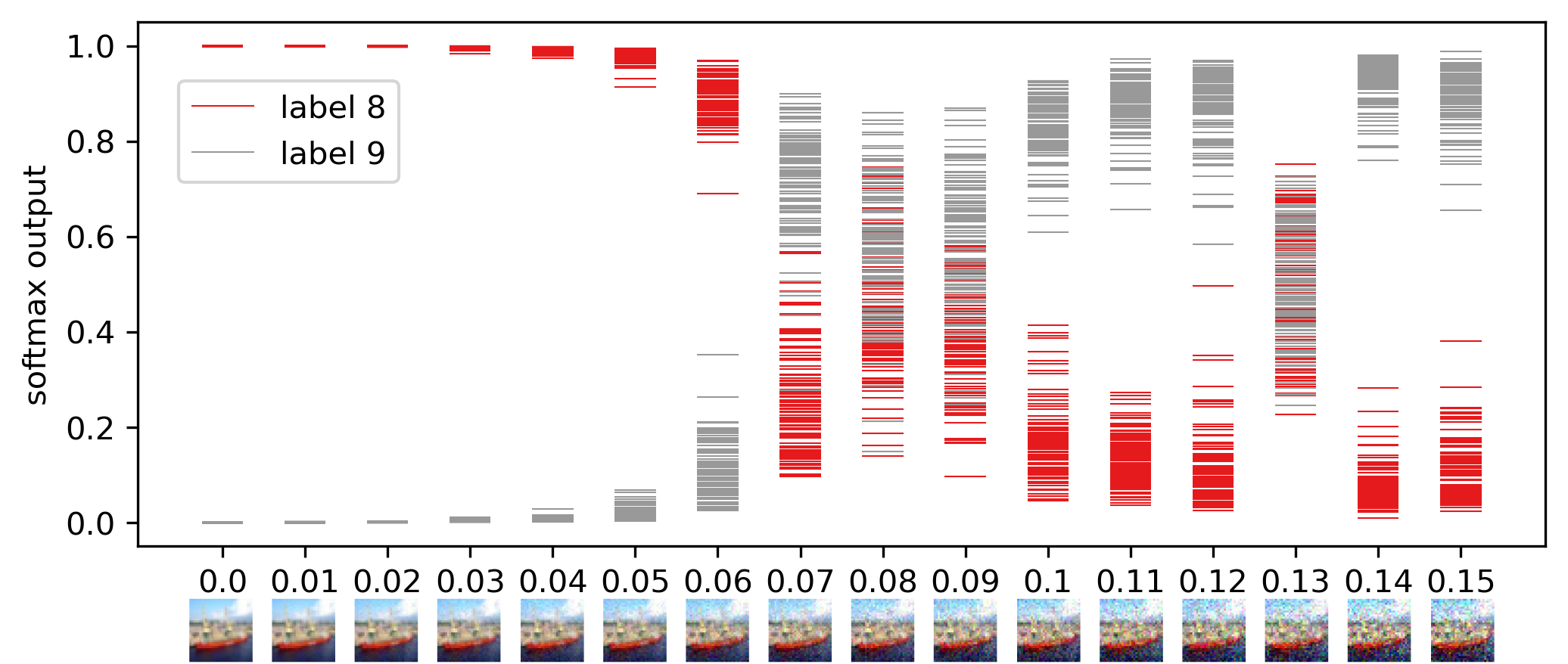}}
\end{minipage}
}
\subfigure[MCNI (learned)]
{\begin{minipage}[b]{0.45\linewidth}
    \centering
    {\includegraphics[width=1\linewidth]{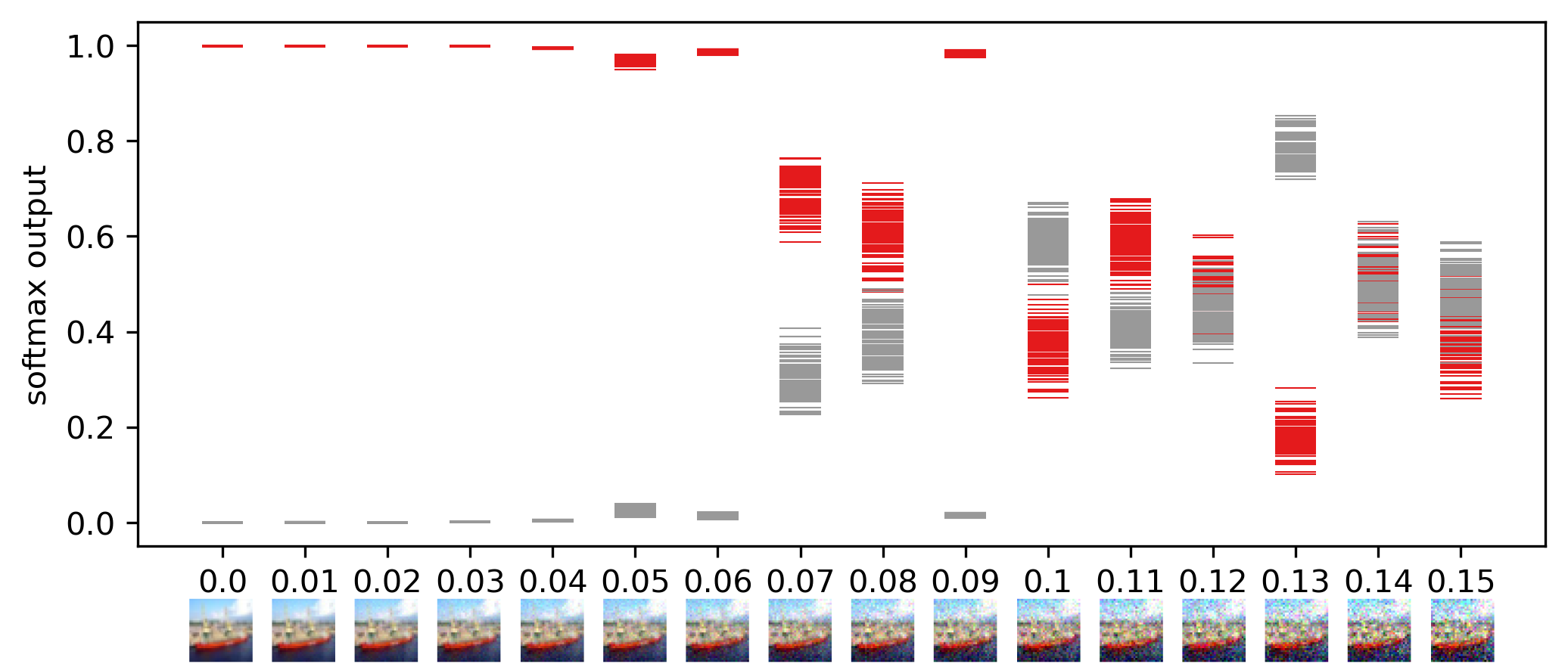}}
\end{minipage}
}
\caption{Softmax outputs from the deterministic NN, MCNI, and MC dropout. The x-axis represents the standard deviation of the Gaussian noise added to the data.}
\label{softmax}
\end{figure*}
\begin{table*}[htbp]
\caption{Test RMSE and MSLL (mean $\pm$ std) of the deterministic NN, MC dropout, and MCNI on the UCI regression datasets. The experiment is repeated five times.}
\label{regression}
\begin{center}
\scalebox{0.9}
{\begin{tabular}{c|cccc|ccc}
   \toprule
   \multirow{2}*{Dataset} & \multicolumn{4}{c}{Test RMSE} & \multicolumn{3}{c}{Test MSLL} \\
   ~ & Deterministic NN & MC dropout & MCNI (fixed) & MCNI (learned) & MC dropout & MCNI (fixed) & MCNI (learned) \\
   \midrule
    Boston Housing & $0.1308\pm 0.0058$ & $0.1340\pm 0.0068$ & $0.1377\pm 0.0105$ & $\boldsymbol{0.1298}\pm 0.0067$ & $0\pm 0.86$ & $-15.53\pm 0.74$ & $\boldsymbol{-66.16}\pm 0.90$ \\
    Concrete Strength & $0.0996\pm 0.0063$ & $0.1003\pm 0.0075$ & $0.1001\pm 0.0068$ & $\boldsymbol{0.0987}\pm 0.0066$ & $0\pm 0.53$ & $-2.17\pm 0.66$ & $\boldsymbol{-43.24}\pm 0.64$ \\
    Energy Efficiency & $0.0102\pm 0.0019$ & $0.0090\pm 0.0030$ & $\boldsymbol{0.0078}\pm 0.0023$ & $0.0097\pm 0.0026$ & $\boldsymbol{0}\pm 0.19$ & $12.87\pm 0.21$ & $22.25\pm 0.16$ \\
    Kin8nm & $0.0757\pm 0.0011$ & $0.0752\pm 0.0018$ & $\boldsymbol{0.0716}\pm 0.0013$ & $0.0738\pm 0.0013$ & $0\pm 1.34$	& $\boldsymbol{-43.74}\pm 1.22$ & $-22.47\pm 1.43$ \\
    Power Plant & $0.0498\pm 0.0003$ & $0.0496\pm 0.0006$ & $\boldsymbol{0.0491}\pm 0.0006$ & $0.0496\pm 0.0008$ & $0\pm 2.32$ & $\boldsymbol{-461.91}\pm 2.43$ & $-221.28\pm 2.33$\\
    Protein Structure & $0.4762\pm 0.0123$ & $0.4649\pm 0.0146$ & $\boldsymbol{0.4458}\pm 0.0139$ & $0.4637\pm 0.0140$ & $0\pm 1.36$ & $-558.31\pm 1.02$ & $\boldsymbol{-952.28}\pm 1.86$ \\
    Wine Quality Red & $0.5974 \pm 0.0108$ & $0.5968\pm 0.0076$ & $\boldsymbol{0.5913}\pm 0.0075$ & $0.5961\pm 0.0068$ & $\boldsymbol{0}\pm 1.66$ & $14.67\pm 1.46$ & $11.86\pm 1.53$\\
    Yacht & $0.0406\pm 0.0010$ & $0.0388\pm 0.0017$ &$ 0.0398\pm 0.0018$ & $\boldsymbol{0.0377}\pm 0.0018$ & $0\pm 1.20$ & $-20.09\pm 1.32$ & $\boldsymbol{-20.25}\pm 1.24$\\
   \bottomrule
\end{tabular}}
\end{center}
\end{table*}
\subsection{Demonstration of Uncertainty on CIFAR10}
We select an image labeled as class 8 (ship) from CIFAR10 as an example, perform 100 forward passes, and plot the softmax outputs as shown in Fig. \ref{softmax}. To compare the robustness of different models, we gradually add Gaussian noise with increasing variance to the image. As the noise increases, the model misclassifies the image as class 9 (truck). It can be observed that for the vanilla model, when the standard deviation of the noise reaches or exceeds 0.06, the model begins to misclassify. For MC Dropout and MCNI, as the noise increases, the softmax outputs become more dispersed, indicating higher uncertainty. For MC Dropout and MCNI (fixed), when the standard deviation of the input noise is greater than or equal to 0.07, the softmax output for class 9 exceeds that of class 8. However, the uncertainty in the softmax output of MCNI (fixed) is higher than that of MC dropout. This suggests that our model is better at signaling a lack of confidence in its predictions. MCNI (learned) appears more robust to noise, as it makes correct predictions until the noise standard deviation reaches 0.1.

\begin{table}[htbp]
\caption{Test accuracy, ECE and Brier score (mean $\pm$ std) of the deterministic NN, MC dropout, and MCNI on CIFAR10. The experiment is repeated five times.}
\label{classification}
\begin{center}
\resizebox{\linewidth}{!}
{\begin{tabular}{cccc}
   \toprule
    Model & Accuracy & ECE & Brier score \\
   \midrule
    Deterministic & $\mathbf{72.05\pm 0.29}$ & $0.0198 \pm 0.003$ & $0.3874 \pm 0.003$ \\
    MC dropout & $70.79\pm 0.71$ & $0.0187 \pm 0.004$ & $0.3885 \pm 0.010$ \\
    MCNI(fixed) & $71.08\pm 0.79$ & $0.0192 \pm 0.005$ & $0.3854 \pm 0.010$ \\
    MCNI(learned) & $71.19 \pm 0.80$ & $\mathbf{0.0143 \pm 0.005}$ & $\mathbf{0.3833 \pm 0.011}$ \\
   \bottomrule
\end{tabular}}
\end{center}
\end{table}
\begin{figure*}[h]
\centering
\subfigure[Boston Housing]
{\begin{minipage}[b]{0.3\linewidth}
    \centering
    {\includegraphics[width=1\linewidth]{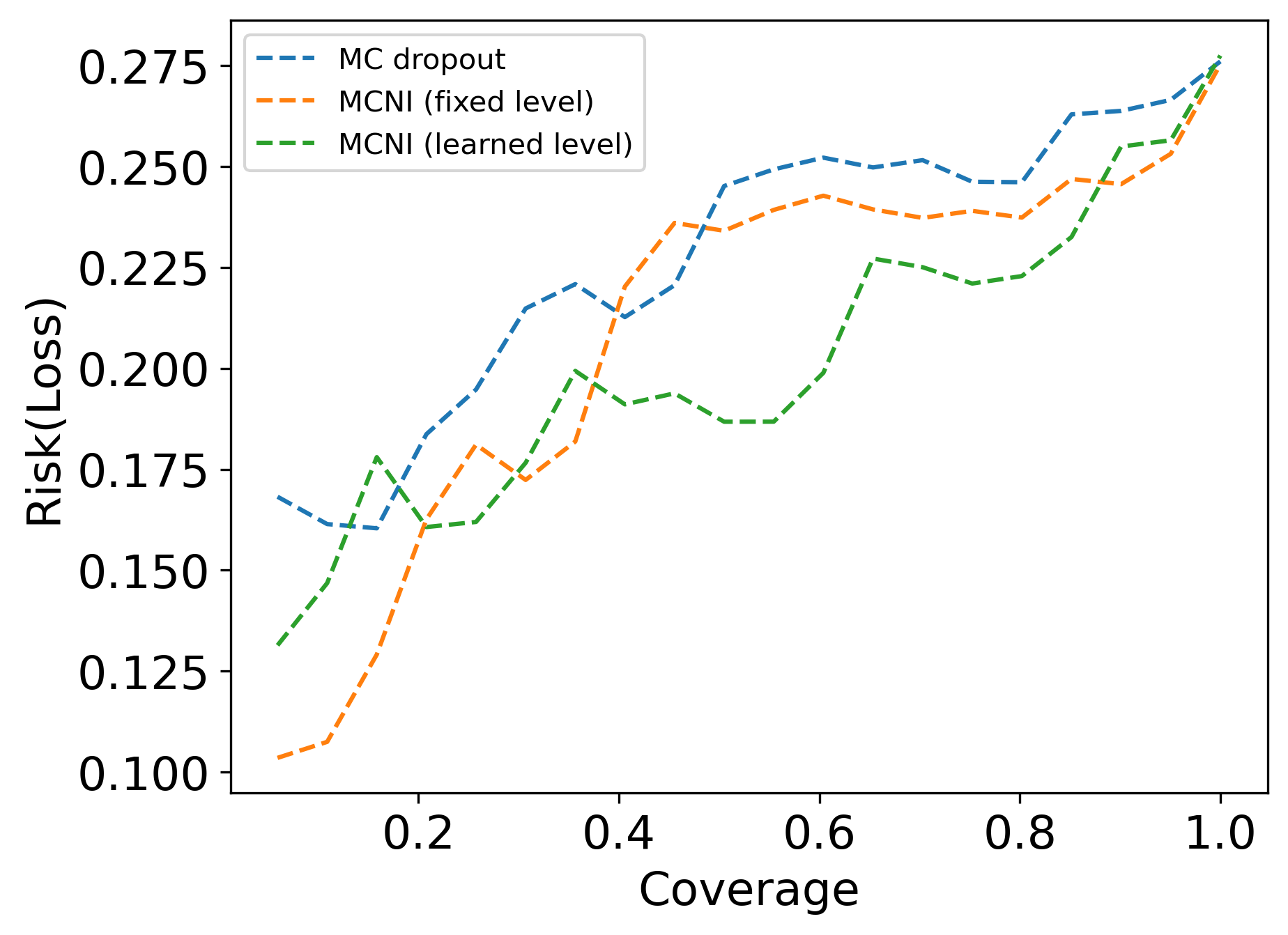}}
\end{minipage}
}
\subfigure[Kin8nm]
{\begin{minipage}[b]{0.3\linewidth}
    \centering
    {\includegraphics[width=1\linewidth]{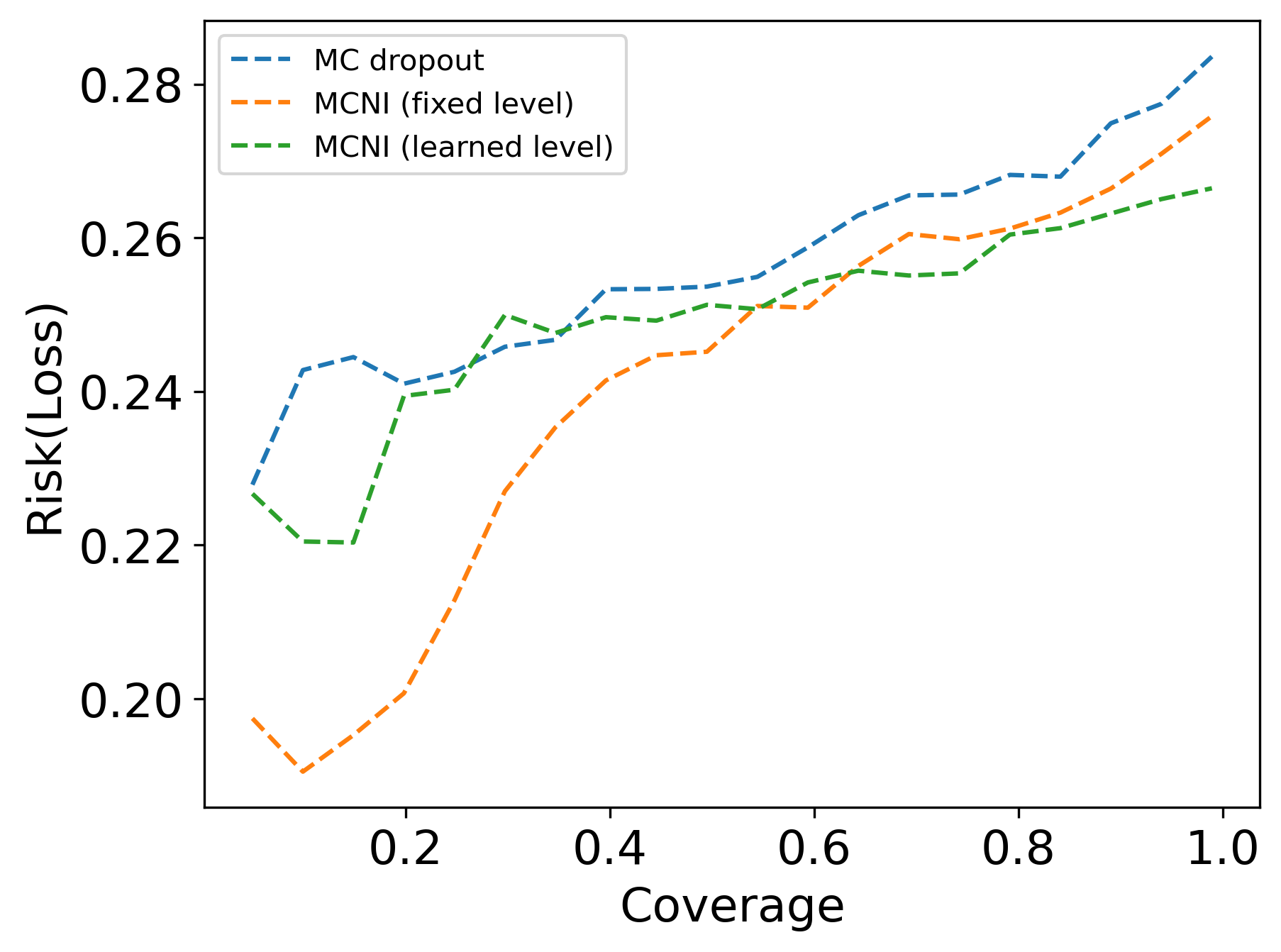}}
\end{minipage}
}
\subfigure[CIFAR10]
{\begin{minipage}[b]{0.3\linewidth}
    \centering
    {\includegraphics[width=1\linewidth]{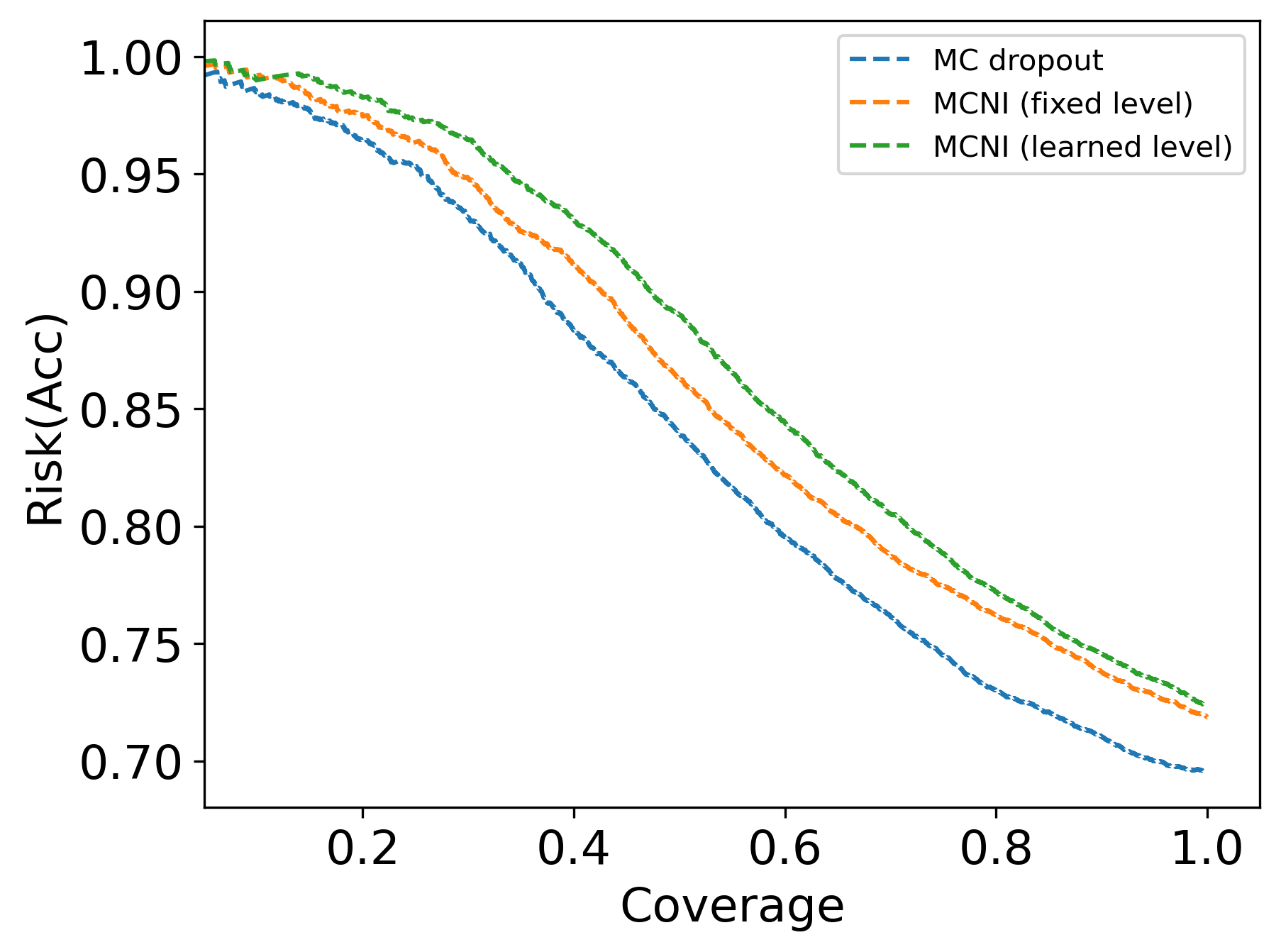}}
\end{minipage}
}
\caption{Risk-Coverage curve on datasets Boston Housing, Kin8nm, and CIFAR10. The risk of the regression task is measured by RMSE and the risk of the classification task is measured by the accuracy.}
\label{risk_coverage}
\end{figure*}
\subsection{Results}
The test RMSE and MSLL for the regression tasks are shown in Table \ref{regression}. On all datasets, MCNI achieves the lowest RMSE, demonstrating stronger predictive performance. MCNI (fixed) outperforms MCNI (learned) on more datasets, possibly due to the additional learnable parameters in MCNI (learned), which may reduce its generalizability. For MSLL, MCNI also outperforms MC Dropout on most datasets. Although it does not perform best on two datasets, the difference is small compared to its better performance on others.

The results of the classification task are shown in Table \ref{classification}. Although the introduction of randomness causes a slight decrease in accuracy relative to the deterministic NN, the proposed MCNI still outperforms the baseline MC dropout. Moreover, MCNI outperforms others in both calibration performance metrics, ECE and Brier score.


We select Boston Housing and Kin8nm for regression tasks, and use CIFAR10 for the classification task, to plot the risk-coverage curve and measure selective performance. The risk-coverage curves are shown in Fig. \ref{risk_coverage}. As the coverage rate increases, the model's risk also rises (RMSE increases or accuracy decreases), indicating that our method effectively identifies high-risk samples. In general, for a given coverage rate, MCNI tends to exhibit lower risk (smaller RMSE or higher accuracy) compared to MC dropout in most scenarios, which is consistent with the conclusion that MCNI has better predictive performance.

\section{Discussion and Conclusion}\label{conclusion}
\subsection{Inference Time Analysis}
Inference is performed on the CIFAR10 test set using the same settings as described in \ref{dataset_model}, with a batch size of 500. The inference time is recorded, and the experiment is repeated 10 times to obtain the average result, which is presented in Table \ref{inference_time}. It can be observed that there is no significant difference in the time taken for MCNI and MC dropout. Furthermore, the increase in time for 10 forward propagations, compared to a single forward propagation, is minimal.

\begin{table}[!htbp]
\caption{Inference time for deterministic NN, MC dropout and MCNI.}
\label{inference_time}
\begin{center}
\begin{tabular}{ccccc}
   \toprule
   \diagbox{Model}{\makecell{\# of forward \\passes}} & 1 & 10 & 50 & 100 \\
   \midrule
    Deterministic & 0.9833s & \textbackslash & \textbackslash & \textbackslash \\
    MC dropout & 1.1432s & 1.2987s & 2.4105s & 4.5809s \\
    MCNI(fixed) & 1.1012s & 1.2587s & 2.4894s & 4.9205s \\
    MCNI(learned) & 1.0724s	& 1.2778s & 2.4655s & 4.7708s \\
   \bottomrule
\end{tabular}
\end{center}
\end{table}

\subsection{Conclusion}
In this paper, we establish theoretically that introducing noise into a neural network's weights is equivalent to Bayesian inference on a deep Gaussian process, by selecting a specific covariance function and variational distributions. Consequently, we propose a Monte Carlo Noise Injection (MCNI) technique, which involves incorporating noise into the parameters during training and executing forward propagation multiple times during inference to evaluate the uncertainty of prediction. Our assessment of the predictive distribution using a toy dataset reveals that our method produces better results compared to MC dropout. Moreover, we evaluate predictive performance, calibration performance and selective performance on both regression and classification tasks using UCI regression datasets and CIFAR10, demonstrating that our method outperforms the baseline model.

Since weight noise injection is commonly used in adversarial training and model uncertainty can also help identify adversarial samples, a potential future research direction could involve integrating the two. Exploring the application of this method in other domains, such as uncertainty quantification in reinforcement learning, could be considered. Our future research will also address non-Gaussian noise injection motivated by the results in \cite{yuan2024robustness}. Further, we will also study noise injection in graphs and consider adaptive signal processing methods for noise robustness \cite{yan2022adaptive, costagli2007image}, and the effect of NN architecture on the noise injection performance \cite{kuo2022neural}.

\bibliography{paper}
\end{document}